\documentclass[twocolumn]{article}

\usepackage{arxiv}

\usepackage[utf8]{inputenc} % allow utf-8 input
\usepackage[T1]{fontenc}    % use 8-bit T1 fonts
\usepackage[english]{babel}

\usepackage{float}
\usepackage{stfloats} % best float around two column text
\usepackage{csquotes} % correct typo for citation, using \enquote{...}

\usepackage{url}            % simple URL typesetting
\usepackage[hidelinks]{hyperref}       % hyperlinks

\usepackage{tabularx}
\usepackage{booktabs}       % professional-quality tables

\usepackage{microtype}      % microtypography
\usepackage{cleveref}       % smart cross-referencing of figures and tables in text, using \cref{fig:example}
\crefname{figure}{Figure}{Figures}
\Crefname{figure}{Figure}{Figures}
\crefname{table}{Table}{Tables}
\Crefname{table}{Table}{Tables}

\usepackage{graphicx}
\usepackage{caption}
\usepackage[authoryear,round]{natbib} % use round but squarish brackets
\usepackage{enumitem} % advanced enumeration

\title{The Prompt Canvas: A Literature-Based Practitioner Guide for Creating Effective Prompts in Large Language Models}
\date{December 6, 2024}
\newif\ifuniqueAffiliation
\ifuniqueAffiliation % Standard variant of author block
\author{Michael~Hewing\\
	FH Münster -- University of Applied Science\\
	\texttt{michael.hewing@fh-muenster.de} \\
	\And
	Vincent~Leinhos\\
	FH Münster -- University of Applied Science\\
	\texttt{vincent.leinhos@fh-muenster.de} \\
}
\else
\usepackage{authblk}

\author{%
	Michael~Hewing%
}
\author{%
	Vincent~Leinhos%
}
\affil{FH Münster -- University of Applied Sciences}
\fi

\renewcommand{\headeright}{}
\renewcommand{\undertitle}{}
\renewcommand{\shorttitle}{}

\hypersetup{
pdftitle={The Prompt Canvas},
pdfauthor={Michael~Hewing, Vincent~Leinhos},
}
\begin{document}
\onecolumn
\maketitle
\begin{abstract}
The rise of large language models (LLMs) has highlighted the importance of prompt engineering as a crucial technique for optimizing model outputs. While experimentation with various prompting methods, such as Few-shot, Chain-of-Thought, and role-based techniques, has yielded promising results, these advancements remain fragmented across academic papers, blog posts and anecdotal experimentation. The lack of a single, unified resource to consolidate the field’s knowledge impedes the progress of both research and practical application. This paper argues for the creation of an overarching framework that synthesizes existing methodologies into a cohesive overview for practitioners. Using a design-based research approach, we present the Prompt Canvas (\cref{fig:thepromptcanvas}), a structured framework resulting from an extensive literature review on prompt engineering that captures current knowledge and expertise. By combining the conceptual foundations and practical strategies identified in prompt engineering, the Prompt Canvas provides a practical approach for leveraging the potential of Large Language Models. It is primarily designed as a learning resource for pupils, students and employees, offering a structured introduction to prompt engineering. This work aims to contribute to the growing discourse on prompt engineering by establishing a unified methodology for researchers and providing guidance for practitioners.
\end{abstract}
\vskip 0.06in
\enlargethispage{6\baselineskip}
\begin{figure}[!htb]
  \centering
  \includegraphics[width=0.78\textwidth]{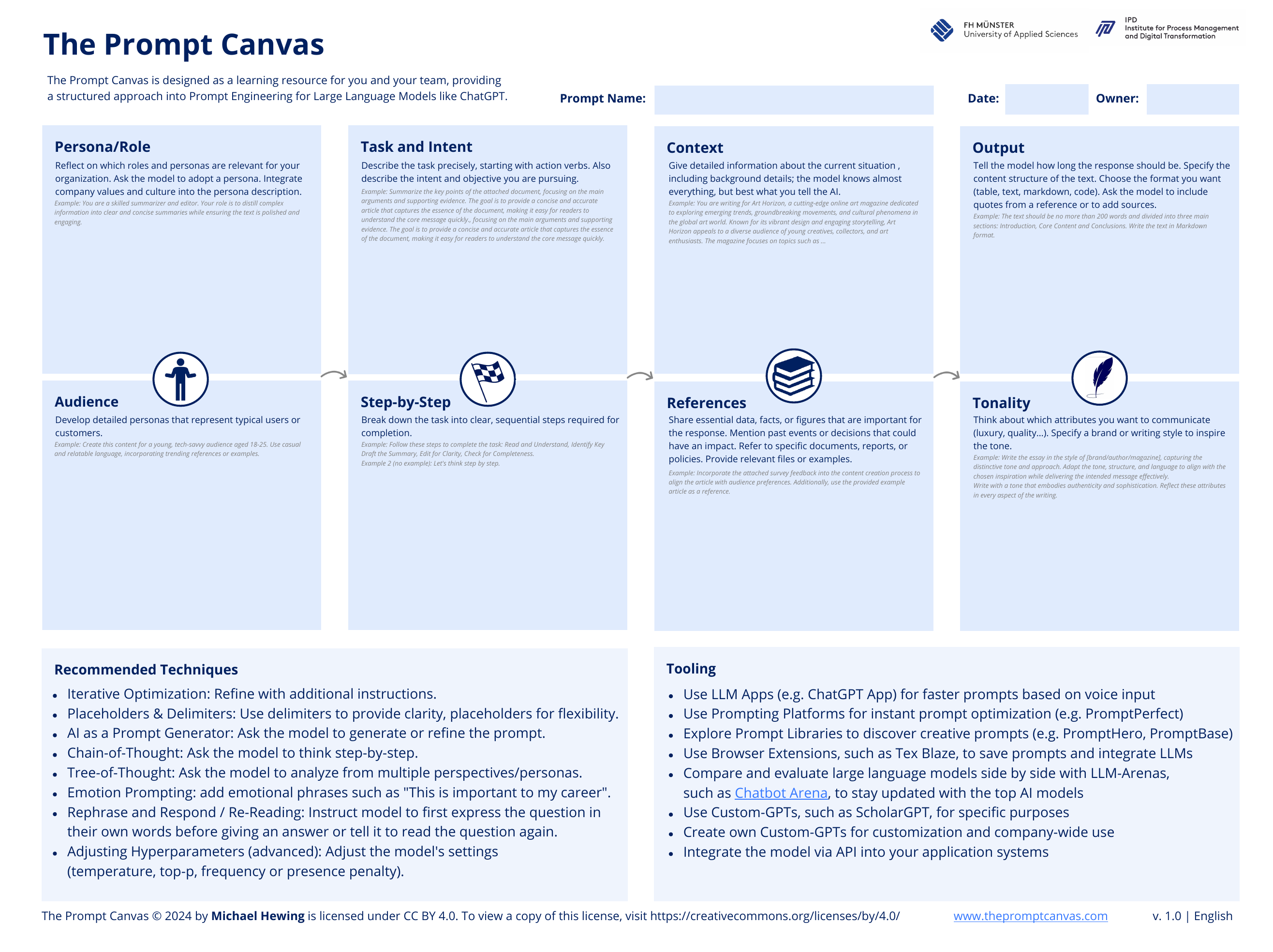}
  \caption{The Prompt Canvas.}
  \label{fig:thepromptcanvas}
\end{figure}
\clearpage
\twocolumn
\vspace*{-0.45in}
\vspace*{-\baselineskip}
\section{Introduction}
With the advances of sophisticated Large Language Models (LLM), the ability to guide these models to generate useful, contextually relevant, and coherent answers has become an essential skill. Prompt engineering refers to the art and science of designing inputs or queries (prompts) that effectively guide LLMs towards desired outputs. \citet[p.~7]{schulhoff_prompt_2024} describe related prompt techniques as a \enquote{blueprint that outlines how to structure a prompt.} This discipline bridges the gap between the user's goals and the model's capabilities, enabling more precise, creative, and domain-specific solutions. Prompt engineering allows users to precisely guide LLMs in generating contextually relevant and task-specific responses.\\
Yet, much of the research and insights into prompt engineering are distributed across disparate sources, such as academic journals, preprints, blogs, and informal discussions on platforms like GitHub, Reddit, or YouTube. Navigating this complex landscape requires not only significant effort, but also a level of expertise that may be inaccessible to practitioners, creating a substantial barrier to entry and hindering the effective application of prompt engineering techniques in practice. With this paper, a canvas-oriented approach is proposed that consolidates current knowledge in the field of prompt engineering into a coherent, visual format. This way, practitioners can implement effective strategies more confidently and with clarity.\\
The second chapter of this paper describes the fragmented state of knowledge in prompt engineering, highlighting the challenges practitioners face in accessing and applying diverse techniques. In the third chapter, a comprehensive review of existing studies and approaches in prompt engineering is presented, showcasing key techniques and patterns in the field. Chapter Four introduces the Prompt Canvas as a structured framework to consolidate and visually represent prompt engineering techniques for better accessibility and practical application. The last chapter provides a conclusion along with constraints and areas for future research.
\section{The Need for an Overview of Prompt Engineering Techniques}
Prompts are vital for LLMs because they serve as the primary mechanism for translating user intentions into actionable outputs. By guiding the model's responses, prompts enable LLMs to perform a wide range of tasks, from creative writing to complex problem-solving, without requiring task-specific retraining. They leverage the pre-trained knowledge embedded in the model, allowing users to adapt LLMs to specific contexts and applications through in-context learning.
\subsection{The Relevance of Prompting in Unsupervised Learning and Transformer Architecture}
Through the concepts of unsupervised learning and basic transformer architecture, the relevance and impact of prompts is evident. Generative AI models, such as LLMs, can be assigned to natural language processing (NLP) in the field of artificial intelligence \citep[p.~560]{braun_can_2024}. In an earlier paradigm of NLP \citep[p.~4]{liu_pretrain_2023}, models were typically trained for specific tasks using supervised learning with manually annotated data. However, this limited a model to its training domain and manual annotation during training was time-consuming and expensive  \citep[p.~1]{radford_improving_2018}. This challenge led to unsupervised learning gaining in importance. According to P. Liu et al., this represents the transition to the current NLP paradigm of \textit{Pre-train, Prompt, Predict}. Large and diverse data sets are used for training. In this way, the model recognizes patterns and aligns parameters within a neural network. By entering a prompt, the model adapts to the corresponding task, which is known as \textit{in-context learning}. This allows the model to be used for a variety of tasks (see \citealp[p.~2]{radford_improving_2018}; \citealp[p.~3]{brown_language_2020}). In addition to unsupervised learning, the transformer architecture, which was published in 2017 by \citeauthor{vaswani_attention_2023} under the title \enquote{Attention Is All You Need,} laid an important foundation for today's LLMs. It enables context to be maintained across long texts \citep[p.~2]{radford_improving_2018}. In June 2018, \citet{openai_improving_2018_blog} stated that their \enquote{\dots approach is a combination of two existing ideas: transformers and unsupervised pre-training.} The abbreviation GPT, Generative Pre-Trained Transformer, reflects this approach. The process from prompt input to output is described in \citet{lo_art_2023} as follows (the process description has been shortened for clarity): First, individual words of the prompt are broken down into tokens. Each token is represented by a vector that conveys its meaning. This representation is referred to as embedding. Self-attention is used to capture the relationship between tokens in the prompt. Finally, based on the previous context and the patterns learned in the training data, next tokens are predicted. Once a token has been selected, it is translated back into a human-readable form. This process is repeated until a termination criterion is reached.\\
These foundational concepts enable LLMs to process diverse and complex tasks without task-specific retraining, relying instead on adaptive responses generated through prompting. Prompt engineering bridges the gap between generalized pre-trained knowledge and specific user needs, functioning as the key mechanism through which the model's potential is harnessed. The transformer’s self-attention mechanism ensures contextual integrity across sequences, while unsupervised learning enables the model to identify and generalize patterns from vast datasets. Together, these innovations allow LLMs to excel in the Pre-train, Prompt, Predict paradigm, making prompt engineering not only a critical aspect of model utility but also a determinant of task-specific success. As AI applications expand across domains, the ability to craft precise and effective prompts will remain central to realizing the full power of these transformative technologies.
\subsection{The Need for a Practitioner-Oriented Overview on Prompt Engineering Techniques}
Prompt engineering is a rapidly evolving field, with techniques such as Few-shot learning, Chain-of-Thought reasoning and iterative feedback loops being developed and refined to solve complex problems. The pace of innovation is driven by a wide range of applications in industries such as healthcare, education and software development, where tailored prompts can significantly improve model performance. A large body of research is investigating the effectiveness of different prompting techniques. However, the current state of knowledge in this area is highly fragmented, posing significant challenges to researchers and practitioners alike. Fragmentation of knowledge refers to the disjointed and inconsistent distribution of information across various sources, often lacking coherence or standardized frameworks. One of the primary challenges of this fragmented knowledge is the absence of a unified framework that consolidates the diverse techniques, methodologies and findings in prompt engineering. Practitioners new to the field face steep learning curves, as they must navigate a scattered and complex body of literature.\\
Yet, as it will be highlighted in the literature review of chapter Three, initial efforts to systematically consolidate these techniques, develop taxonomies and establish a shared vocabulary are emerging. These publications structure current knowledge into schemes and patterns. While they provide in-depth analyses and valuable structures, they often lack accessibility for practitioners seeking practical solutions and actionable insights. This gap from research advancements to practical application highlights a pressing need for bridging between academic research and real-world use. Addressing these challenges will ensure that the benefits of prompt engineering are more widely realized, enabling its application to expand further across industries and domains.
\subsection{Canvas for Visualization}
The field of prompt engineering involves a dynamic and multifaceted interplay of strategies, methodologies, and considerations, making it challenging to present in a way that is both comprehensive and accessible. The canvas model promotes visual thinking and has been widely adopted in fields such as business strategy (\citealt{Osterwalder2010}, \citealt{Pichler2016}), teamwork \citep{IvanovVoloshchuk2015}, startups \citep{Maurya2012}, research \citep{FountainInstitute2020} and design thinking \citep{IBM2016}, where it has proven to be an effective way to organize and communicate complex processes. A canvas simplifies complexity by visually organizing aspects of relevance into defined sections, allowing users to see the relationships and workflows at a glance. It promotes a holistic view of the process in one unified space. Also, the collaborative nature of a canvas facilitates communication and alignment among team members with varying levels of expertise.\\
By applying this proven framework to prompt engineering and making the transition to this visual representation more intuitive, practitioners can leverage prompt techniques and patterns. Practitioners can quickly grasp the key elements and a workflow, reducing barriers to entry and enabling more effective application of the techniques.
\section{Identifying Common Techniques Through a Systematic Literature Review}
In order to obtain a comprehensive overview of the current state of techniques in the field of prompt engineering, a systematic literature review (SLR) has been carried out. Such a systematic approach provides transparency in the selection of databases, search terms, as well as inclusion and exclusion criteria. After the literature search and selection, included literature will be analyzed and consolidated.
\subsection{Literature Search and Selection}
\begin{figure*}[!htb]
  \centering
  \includegraphics[width=0.6\textwidth]{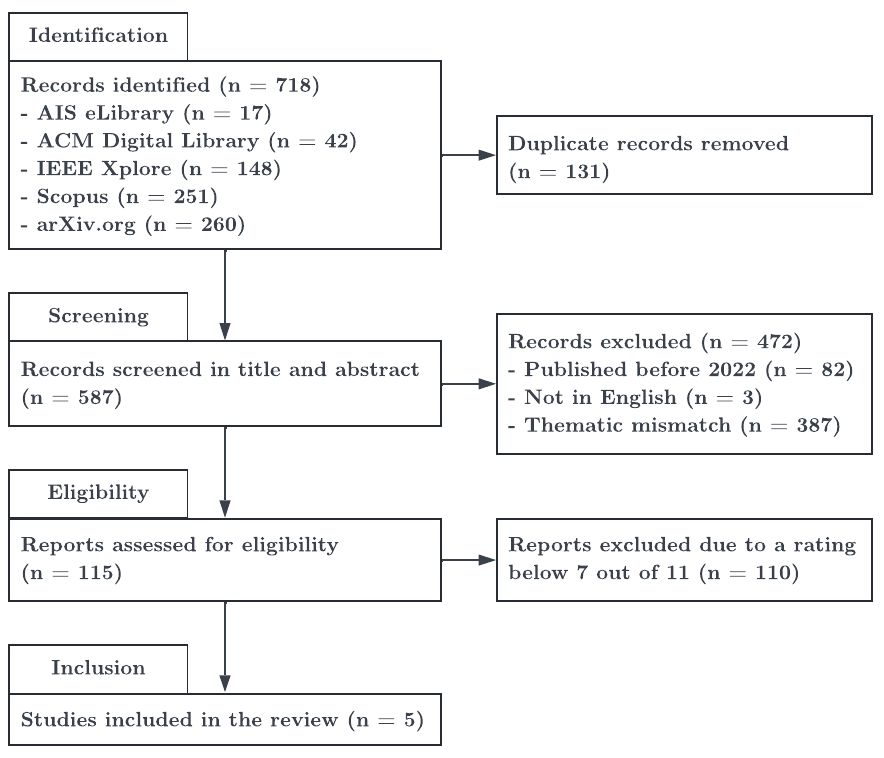}
  \caption{PRISMA procedure, based on \protect\citep[p.~5]{prisma2020}.}
  \label{fig:slr_prisma_flow}
\end{figure*}

The literature search process primarily adheres to the framework outlined by \citet[pp.~8--11]{brocke2009}. For the subsequent selection of sources, the methodology is based on the Preferred Reporting Items for Systematic reviews and Meta-Analyses (PRISMA) guidelines (cf. Page et al., 2021). Vom Brocke et al. (2009) outline the systematic literature review (SLR) process in five distinct phases. The process begins with defining the scope of the literature search (Phase 1) and creating a preliminary concept map (Phase 2) to guide the review. This is followed by the execution of the actual literature search (Phase 3). The later stages involve the analysis and synthesis of the included literature (Phase 4) and a discussion of the findings along with their limitations (Phase 5). The last phase we integrated into the section on limitations at the end of this paper. Vom Brocke et al. (2009) emphasize the first three phases in their work. With the literature research, the following research question shall be addressed:\\
\textit{What is the current state of techniques and methods in the field of prompt engineering, especially in text-to-text modalities?}\\
To establish the framework for the literature search in phase one, \citet{brocke2009} draw on Cooper’s taxonomy (\citeyear[pp.~107--112]{cooper1988}). Cooper identifies six key characteristics for classifying literature searches: focus, goal, perspective, coverage, organization and audience. These characteristics provide a structured approach to defining the purpose and scope of a literature review. \cref{tab:cooper1988_taxonomy_lr_characteristics} offers a detailed overview of how these classifications align with the specific intentions of this SLR, ensuring a systematic and targeted review process.
% tab:cooper1988_taxonomy_lr_characteristics
\begin{table}[!htb]
\caption{Characteristics according to Cooper (\protect\citeyear[pp.~107--112]{cooper1988}) applied to this SLR.}
\vspace{0.1in}
\begin{tabularx}{\columnwidth}{lX}
\toprule
\textbf{Characteristic} & \textbf{Category}\\
\midrule
\textbf{Focus} & Research outcomes, Practices or applications\\
\addlinespace
\textbf{Goal} & Integration or synthesis\\
\addlinespace
\textbf{Perspective} & Neutral representation \\
\addlinespace
\textbf{Coverage} & Exhaustive coverage with selective citation\\
\addlinespace
\textbf{Organization} & Conceptual (thematically organized)\\
\addlinespace
\textbf{Audience} & Users of LLMs (private and business use)\\
\bottomrule
\end{tabularx}
\label{tab:cooper1988_taxonomy_lr_characteristics}
\end{table}
The second phase involves elaboration using concept mapping. For this purpose, terms are selected that are expected to lead to relevant and comprehensive results in the subsequent database search. To keep the literature review as inclusive as possible, only terms directly related to prompt engineering were included: \textit{prompt engineering}, \textit{prompt techniques}, \textit{prompt designs}, \textit{prompt patterns}, \textit{prompt strategies}, \textit{prompt methods}. Further related concepts such as \textit{LLMs} or \textit{generative AI} have not been considered because they might broaden the scope too much.\\
According to vom Brocke et al. (2009), the third phase is divided into several steps. The first step is to identify and select qualitative sources for the literature review. The \enquote{VHB Publication Media Rating 2024} for the section Information Systems is an established reference for quality and impact of sources \citep{VHB2024}. Journals with a VHB rating of B or higher and a potential focus on AI were preselected. This selection was made by manually reviewing the short descriptions of the respective journals and evaluating their relevance with the assistance of generative AI. Prompt used in \textit{GPT-4o} on October 4, 2024: \enquote{Evaluate which of the following journals could contain articles relevant to the topic of prompt engineering.} Based on the manual and AI-supported selection, the following journals should at least be included in the database set for this literature search: \textit{Nature Machine Intelligence}, \textit{ACM Transactions on Computer-Human Interaction (TOCHI)}, \textit{Artificial Intelligence (AIJ)}, \textit{IEEE Transactions on Knowledge and Data Engineering}, \textit{ACM Transactions on Interactive Intelligent Systems (TiiS)}.\\
It is understood that conferences play an important role in the field of generative AI, as the timely exchange of new approaches is fundamental. Taking into account a VHB rating of B or higher and an evaluation of thematic relevance, the following conferences should also be included in the database: International Conference on Information Systems (ICIS), European Conference on Information Systems (ECIS), Hawaii International Conference on System Sciences (HICSS), International Conference on Machine Learning (ICML), Association for the Advancement of Artificial Intelligence (AAAI), International Joint Conference on Artificial Intelligence (IJCAI), ACM Conference on Human Factors in Computing Systems (CHI), Conference on Neural Information Processing Systems (NeurIPS).\\
The next step within the third phase, the selection of databases and search terms, is to search for peer-reviewed SLRs on the topic of prompt engineering to identify databases relevant to the subject. The search was conducted on Scopus on the 10th of August and resulted in five hits (\cref{tab:slr_first_search_scopus}). The title, abstract and keywords were searched using the following search term:\\
\texttt{( TITLE-ABS-KEY ( \char34 prompt engineering\char34 ) OR TITLE-ABS-KEY ( \char34prompt-engineering\char34 ) AND TITLE-ABS-KEY ( \char34 systematic literature review\char34 ) OR TITLE-ABS-KEY ( \char34 PRISMA\char34 ) )}
% tab:slr_first_search_scopus
\begin{table}[!htb]
\caption{Search results for systematic literature reviews in the field of prompt engineering (performed in Scopus on August 10, 2024).}
\vspace{0.1in}
\begin{tabularx}{\columnwidth}{p{1.8cm}p{1.8cm}X}
\toprule
\textbf{Reference} & \textbf{Domain} & \textbf{Databases}\\
\midrule
\citet{sasaki2024} & 
Programming &
Google Scholar; arXiv, ACM Digital Library, IEEE Xplore\\
\addlinespace
\citet{moglia2024} & 
Medicine &
PubMed, Web of Science, Scopus, arXiv\\
\addlinespace
\citet{han2023} & 
Economy &
JSTOR, ProQuest, ScienceDirect, Web of Science, Google Scholar\\
\addlinespace
\citet{watson2023} & 
Machine Learning &
Scopus, IEEE Xplore, ScienceDirect, Elicit, WorldCat, Google Scholar, arXiv\\
\bottomrule
\end{tabularx}
\label{tab:slr_first_search_scopus}
\end{table}
These publications focus on specific application areas, such as programming, medicine, economics, or machine learning, making it challenging to generalize their insights for broader practical use by practitioners. Yet, similarities in the database selection were recognized. All hits either use arXiv.org directly as a database or cite a large number of their sources from that website. Documents on arXiv.org are largely not peer-reviewed, but on the other hand enable the publication of current research. In the \enquote{Systematic Literature Review of Prompt Engineering Patterns in Software Engineering,} \citet[p.~671]{sasaki2024} transparently state that most of their cited sources are not peer-reviewed, but that it is important to include them because prompt engineering is a rapidly changing field. Another SLR from the preliminary research states that many current articles can only be obtained through arXiv.org and an increasing number of research groups are publishing their work on arXiv \citep[p.~41]{moglia2024}. Based on these findings and the previously identified journals, databases and search terms were defined (see \cref{tab:database_and_keyterms} in Appendix). On the one hand, those qualitative journals and conferences should be considered (accessible via AIS eLibrary, IEEE Xplore, ACM Digital Library), while at the same time fully including interdisciplinary areas (through Scopus), as well as current – albeit largely non-peer-reviewed – articles (from arXiv).\\
The selected search terms result from concept mapping and iterative testing of keywords. It was found that many articles only contained the term prompt engineering in the abstract, but their research focus was in a different area. This could be explained by the fact that prompt engineering can play an indirect role in many areas of application. Since it is assumed that articles with a primary focus on prompt engineering also contain this term in the title, only the title was searched. The following search term with variations was formed:\\
\texttt{TITLE(\char34 prompt-engineering\char34\ OR \char34 prompt engineering\char34\ OR \char34 prompt techniques\char34\ OR \char34 prompt designs\char34\ OR \char34 prompt design\char34\ OR \char34 prompt patterns\char34\ OR \char34 prompt pattern\char34\ OR \char34 prompt strategies\char34\ OR \char34 prompt strategy\char34\ OR \char34 prompt methods\char34 )}\\
The search was carried out on October 4, 2024 in the respective databases according to the PRISMA procedure \citep{prisma2020} (see \cref{fig:slr_prisma_flow}) and documented with the literature management program Zotero. 718 hits were identified for all databases. Of these, 131 hits were identified as duplicates and excluded accordingly. In the next step, 587 hits were checked for suitability with regard to their title and abstract. Previously, articles that were published before 2022 or were not written in English were excluded. In the third step, 115 full-text articles were checked for suitability.
% fig:slr_prisma_flow
Since articles from arXiv.org may not contain peer-reviewed articles, but at the same time are often highly relevant, an evaluation system was created to evaluate articles from all databases holistically according to thematic suitability, quality and actuality. The thematic suitability was weighted most heavily, while the quality of articles was assessed using two criteria to ensure a comprehensive evaluation. First, we prioritized publications that include a literature review process, assigning higher scores to systematic literature reviews (SLRs) and lower scores to less detailed reviews. This is of importance as we want to consolidate the knowledge in this field. Second, the evaluation incorporated the VHB rating (Verband der Hochschullehrerinnen und Hochschullehrer für Betriebswirtschaft e.V.), with higher scores. The evaluation criteria are defined in \cref{tab:slr_rating}.
% tab:slr_rating
\begin{table}[!htb]
\caption{Criteria for evaluating articles in full text.}
\vspace{0.1in}
\begin{tabularx}{\columnwidth}{lX}
\toprule
\textbf{Criterion} & \textbf{Explanation}\\
\midrule
\textbf{Topic} & \textbf{Is the full text of the article relevant to answering the research question?}\\
& 4~=~very relevant\\
& 3~=~relevant\\
& 2~=~somewhat relevant\\
& 1~=~less relevant\\
& 0~=~not relevant\\
\midrule
\textbf{Quality} & \textbf{(1) How transparent is the literature research process of that article?}\\
& 2~=~very transparent (SLR)\\
& 1~=~present (LR)\\
& 0~=~not transparent\\
& \textbf{(2) Does a VHB rating} \citep{VHB2024} \textbf{exist for this article?}\\
& 3~=~A+\\
& 2~=~A\\
& 1~=~B\\
& 0~=~C\\
& 0~=~D or not available\\
\midrule
\textbf{Actuality} & \textbf{When was the article published?}\\
& 2~=~2024\\
& 1~=~2023\\
& 0~=~2022 or before\\
\bottomrule
\end{tabularx}
\label{tab:slr_rating}
\end{table}
Articles scoring fewer than seven points were excluded from the primary selection. However, the articles below this threshold, especially those with a score of six points, were reviewed as supplementary sources. These also include the four SLR articles from the previous database selection.\\
There has been a significant increase in the number of articles published in recent years that can be assigned to the field of prompt engineering based on their title or abstract. Of the 115 articles that were checked for their suitability in full text in the fourth step, 63 articles were published in the year 2024 to date (up to October 4, 2024), 44 articles were published in 2023 and eight articles in 2022. Many articles were related to fine-tuning models, which was not relevant to the research question, as it requires specialized technical knowledge. Ultimately, five articles met the inclusion criteria, demonstrating relevance, quality and alignment with the research question. These five articles are summarized in \cref{tab:slr_results_rating7_numbering}.

% tab:slr_results_rating7_numbering
\begin{table*}[!hb]
\caption{Final inclusion of SLR articles.}
%\vspace{0.1in}
\begin{tabularx}{\textwidth}{llX}
\toprule
\textbf{No.} & \textbf{Reference} & \textbf{Title}\\
\midrule
1 & \citet{braun_can_2024} & Can (A)I Have a Word with You? A Taxonomy on the Design Dimensions of AI Prompts\\
\addlinespace
2 & \citet{schulhoff_prompt_2024} & The Prompt Report: A Systematic Survey of Prompting Techniques\\
\addlinespace
3 & \citet{sahoo_systematic_2024} & A Systematic Survey of Prompt Engineering in Large Language Models: Techniques and Applications\\
\addlinespace
4 & \citet{sasson_lazovsky_art_2024} & The Art of Creative Inquiry—From Question Asking to Prompt Engineering\\
\addlinespace
5 & \citet{white_prompt_2023} & A Prompt Pattern Catalog to Enhance Prompt Engineering with ChatGPT\\
\bottomrule
\end{tabularx}
\label{tab:slr_results_rating7_numbering}
\end{table*}
\begin{table*}[!hbt]
\vspace{0.2in}
\caption{Articles that score six points.}
% tab:slr_results_rating6_numbering
\begin{tabularx}{\textwidth}{llX}
\toprule
\textbf{No.} & \textbf{Reference} & \textbf{Title}\\
\midrule
1 & \citet{bhandari_survey_2024} & A Survey on Prompting Techniques in LLMs\\
\addlinespace
2& \citet{bozkurt_tell_2024} & Tell Me Your Prompts and I Will Make Them True: The Alchemy of Prompt Engineering and Generative AI\\
\addlinespace
3 & \citet{Chen2024} & Unleashing the potential of prompt engineering in Large Language Models: a comprehensive review\\
\addlinespace
4 & \citet{chong_prompting_2024} & Prompting for products: Investigating design space exploration strategies for text-to-image generative models\\
\addlinespace
5 & \citet{fagbohun_empirical_2024} & An Empirical Categorization of Prompting Techniques for Large Language Models: A Practitioner's Guide\\
\addlinespace
6 & \citet{garg_analyzing_2024} & Analyzing the Role of Generative AI in Fostering Self-directed Learning Through Structured Prompt Engineering\\
\addlinespace
7 & \citet{hill_prompt_2024} & Prompt Engineering Principles for Generative AI Use in Extension\\
\addlinespace
8 & \citet{korzynski_artificial_2023} & Artificial intelligence prompt engineering as a new digital competence: Analysis of generative AI technologies such as ChatGPT\\
\addlinespace
9 & \citet{liu_design_2022} & Design Guidelines for Prompt Engineering Text-to-Image Generative Models\\
\addlinespace
10 & \citet{sasaki2024} & Systematic Literature Review of Prompt Engineering Patterns in Software Engineering\\
\addlinespace
11 & \citet{schmidt_towards_2024} & Towards a Catalog of Prompt Patterns to Enhance the Discipline of Prompt Engineering\\
\addlinespace
12 & \citet{siino_gpt_2024} & GPT Hallucination Detection Through Prompt Engineering\\
\addlinespace
13 & \citet{tolzin_worked_2024} & Worked Examples to Facilitate the Development of Prompt Engineering Skills\\
\addlinespace
14 & \citet{tony_prompting_2024} & Prompting Techniques for Secure Code Generation: A Systematic Investigation\\
\addlinespace
15 & \citet{vatsal_survey_2024} & A Survey of Prompt Engineering Methods in Large Language Models for Different NLP Tasks\\
\addlinespace
16 & \citet{wang_review_2023} & Review of large vision models and visual prompt engineering\\
\addlinespace
17 & \citet{wang_prompt_2024} & Prompt engineering in consistency and reliability with the evidence-based guideline for LLMs\\
\addlinespace
18 & \citet{ye_prompt_2024} & Prompt Engineering a Prompt Engineer\\
\bottomrule
\end{tabularx}
\label{tab:slr_results_rating6_numbering}
\end{table*}
18 articles, each with six points, were reviewed and included in the analysis as supplementary sources (\cref{tab:slr_results_rating6_numbering}), but they are not described in as much detail as those with seven points. The most common aspects have been already covered by the extensive reviews of the articles listed above.
\subsection{Analysis and Results}
In "Can (A)I Have a Word with You? A Taxonomy on the Design Dimensions of AI Prompts", \citet{braun_can_2024} develop a taxonomy for the design of prompts for different modalities, such as text-to-text and text-to-image.\\
"The Prompt Report" by \citep{schulhoff_prompt_2024} can be considered the most comprehensive article of the included literature. It considers prompting techniques for the text-to-text modality and gives an insight into other modalities, such as text-to-visuals. At the same time, that article uses the PRISMA approach within a systematic literature review, which increases the transparency of the selected prompting techniques. Prompting techniques are categorized by modality and prompting category.\\ 
"A Systematic Survey of Prompt Engineering in Large Language Models: Techniques and Applications" by \citet{sahoo_systematic_2024} is similarly comprehensive. It classifies the prompting techniques according to application area. Schulhoff et al. and Sahoo et al. present a total of 108 different prompting techniques.\\
"The Art of Creative Inquiry-From Question Asking to Prompt Engineering" by Sasson Lazovsky et al. offers a perspective on the similarities between question formulation and prompt engineering. The article shows which characteristics are important in the interaction between humans and generative AI.\\ 
In "A Prompt Pattern Catalog to Enhance Prompt Engineering with ChatGPT", \citet{white_prompt_2023} present prompt patterns using practical examples for software development. However, according to the authors, these can be transferred to other areas.\\
In order to make a selection from the multitude of prompting techniques, those that were presented in several included articles were prioritized. If a prompting technique includes adapted variants, the parent prompting technique is presented first, followed by possible adaptations.
% fig:braun_characteristics
\begin{figure*}[!htb]
  \centering
  \includegraphics[width=0.9\textwidth]{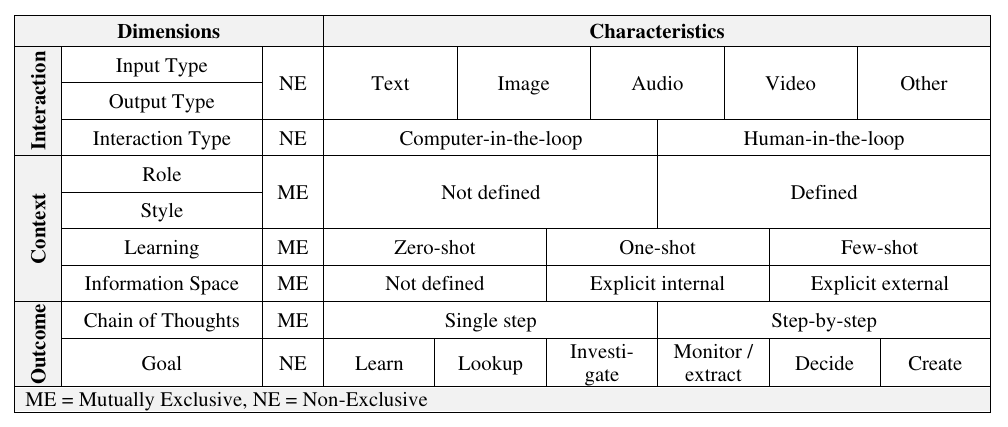}
  \caption{Prompt dimensions and characteristics \protect\cite[p.~563]{braun_can_2024}.}
  \label{fig:braun_characteristics}
\end{figure*}
\citet{braun_can_2024} classified nine dimensions and three meta-dimensions that should be considered when creating prompts (\cref{fig:braun_characteristics}). Firstly, the \textit{interaction} between the LLM and user, which, depending on the prompt, can be seen as \textit{computer-in-the-loop} or \textit{human-in-the-loop} (HITL). An example of HITL would be if the user asks the LLM, in addition to the main instruction, to pose follow-up questions that the user then answers \citep[pp.~561, 565]{braun_can_2024}. In this case, the user would take on a more active role. According to Braun et al., \textit{input} and \textit{output types} such as text, image, audio, video and others are part of the interaction meta-dimension. \textit{Context} is defined as the second meta-dimension. This consists of the \textit{learning} dimension, which is divided into \textit{Zero-shot}, \textit{One-shot} and \textit{Few-shot}. In addition to Braun et al., three of the included articles also identified these as superordinate prompting techniques. As already presented in Chapter Two, an LLM can adapt to new tasks (in-context learning), even if it has not been explicitly trained for that task (\citealp[pp.~563--564]{braun_can_2024}; cf. \citealp{radford_improving_2018}, \citealp{brown_language_2020}). The addition of examples in a prompt is referred to as Few-shot and One-shot in the case of an explicit example. As Brown et al. showed in their article on \textit{GPT-3}, the use of Few-shot can increase the accuracy of the output compared to Zero-shot (see \citealp
{radford_better_2019}), where no examples are provided. In addition, the behavior of an LLM can be adapted to a specific context by assigning a \textit{role} in a prompt (\citealp[p.~564]{braun_can_2024}; cf. \citealp[p.~7]{white_prompt_2023}). By setting a \textit{style}, the output can be adapted more generally \citep[p.~564]{braun_can_2024}. Braun et al. go on to define the information space dimension for the context meta-dimension. The authors differentiate between whether additional information is provided internally - directly in the prompt - or externally - by \textit{agents} that use search engines, for example. If no additional context is provided, the output is based exclusively on the training data of the LLM \citep[p.~564]{braun_can_2024}. Braun et al. define the third and final meta-dimension as the \textit{outcome} to be achieved by adapting a prompt \citep[p.~564]{braun_can_2024}. The authors classify \textit{Chain-of-Thought (CoT)} for this purpose. This was also identified as a superior prompting technique in the articles by \citet{schulhoff_prompt_2024} and \citet{sahoo_systematic_2024}, which refer to the article by \citet{wei_chain_thought_2023}. CoT is designed for tasks that require complex understanding. This is also referred to as reasoning in various included articles. CoT breaks down a problem into smaller steps, solves them and then provides a final answer. This approach aims to provide the user with a clear and understandable result by having the LLM explain the process it uses to generate its output \citep[p.~3]{wei_chain_thought_2023}. Finally, Braun et al. (\citeyear[p.~565]{braun_can_2024}) classified the following goals of a prompt within the meta-dimension result: \textit{learn}, \textit{lookup}, \textit{investigate}, \textit{monitor/extract}, \textit{decide}, \textit{create}. This last dimension concludes Braun et al.'s taxonomy.\\
The article \enquote{A Prompt Pattern Catalog to Enhance Prompt Engineering with ChatGPT} by \citet{white_prompt_2023} makes a further conceptual contribution, which however focuses on concrete prompt structures. White et al. present patterns – so-called prompt patterns, which generally structure prompts for frequently occurring problems in a task area. With the help of such a prompt pattern, prompts can then be formulated for a specific task. This approach not only saves time, but also ensures compliance with proven standards. White et al. present Prompt patterns as examples for the area of software development. Following our own analysis of the prompt patterns presented by White et al., similarities between them were analyzed and classified. These are presented in \cref{tab:white_prompt_take_aways} using examples from \citet{white_prompt_2023}. These observations can be seen as an extension of the previously presented prompt dimensions by \citet[p.~563]{braun_can_2024}.
% tab:white_prompt_take_aways
\begin{table*}[!ht]
\caption{Exemplified prompt patterns from \protect\citet{white_prompt_2023}.}
%\vspace{0.1in}
\begin{tabularx}{\textwidth}{lX}
\toprule
\textbf{Observation} & \textbf{Example}\\
\midrule
\textbf{Scope} & \enquote{Within scope X}\\
\addlinespace
\textbf{Task/Goal} & \enquote{Create a game\dots}\\ & \enquote{I would like to achieve X}\\
\addlinespace
\textbf{Context} & \enquote{When I say X, I mean\dots} \\ & \enquote{Consider Y} \\ & \enquote{Ignore Z}\\
\addlinespace
\textbf{Procedure} & \enquote{When you are asked a question, follow these rules\dots}\\ & \enquote{Explain the reasoning and assumptions behind your answer}\\
\addlinespace
\textbf{Role} & \enquote{Act as persona X. Provide outputs that persona X would create}\\
\addlinespace
\textbf{Output} & \enquote{Please preserve the formatting and overall template that I provide}\\
\addlinespace
\textbf{Termination condition} & \enquote{You should ask questions until this condition is met or to achieve this goal}\\
\bottomrule
\end{tabularx}
\label{tab:white_prompt_take_aways}
\end{table*}
\\
In addition to the selection of a prompt structure and the appropriate selection of prompting techniques, the formulation of questions is essential in order to be able to interact effectively with the generative AI as a user. This requires similar skills to those required for asking interpersonal questions \citep[pp.~7--9]{sasson_lazovsky_art_2024}. Sasson Lazovsky et al. identified the following seven common key skills: \textit{Creativity}, \textit{Clarity and Precision}, \textit{Adaptability}, \textit{Critical Thinking}, \textit{Empathy}, \textit{Cognitive Flexibility}, \textit{Goal Orientation}. These are described in \cref{tab:key_skills_pe}.
% tab:key_skills_pe
\begin{table*}[!ht]
\caption{Prompt Engineering Skills, \protect\citet{sasson_lazovsky_art_2024}.}
%\vspace{0.1in}
\begin{tabularx}{\textwidth}{lX}
\toprule
\textbf{Capability} & \textbf{Applied to the area of prompt engineering}\\
\midrule
\textbf{Creativity} & Designing prompts that elicit desired and insightful responses from AI\\
\addlinespace
\textbf{Clarity and precision} & Conveying instructions precisely to minimize misunderstandings\\
\addlinespace
\textbf{Adaptability} & Tailoring prompts to the task and language model capabilities\\
\addlinespace
\textbf{Critical thinking} & Considering potential outcomes and responses for meaningful interactions\\
\addlinespace
\textbf{Empathy} & Optimizing language model responses through empathetic consideration\\
\addlinespace
\textbf{Cognitive flexibility} & Iterating with various prompts to optimize results\\
\addlinespace
\textbf{Goal orientation} & Eliciting specific responses that align with the intended purpose\\
\bottomrule
\end{tabularx}
\label{tab:key_skills_pe}
\end{table*}
Subsequently, prompting techniques from previously outlined areas such as Zero-shot, Few-shot and Chain-of-Thought will be explored in greater depth and subdivided into further potential areas. The following prompting techniques are taken from the synthesis by  \citet[pp.~8--18]{schulhoff_prompt_2024} and \citet[pp.~2--7]{sahoo_systematic_2024}, who present several prompting techniques and refer to corresponding articles. There are often a large number of articles that adapt a prompting technique for new purposes. Therefore, reference is always made to the original article, unless an adapted variant is presented.\\
Besides Role and Style Prompting, \textit{Emotion Prompting} adds emotional phrases such as \enquote{This is very important to my career} to the end of a prompt \citep[p.~2]{li_large_2023}.
Another area of prompting techniques can be divided into \textit{Rephrase and Re-read}. \textit{Rephrase and Respond} (RaR) instructs an LLM to first express the question in its own words before giving an answer \citep[pp.~9--10]{deng_rephrase_2024}. \textit{Re-reading} (RE2) tells an LLM to read the question again. This can increase performance in the area of reasoning \citep{xu_re-reading_2024}.\\
Prompting techniques that focus on a step-by-step approach can be assigned to the prompting area of Chain-of-Thought. \textit{Chain-of-Thought Zero-shot} adds \enquote{Let's think step by step} at the beginning of a prompt \citep[p.~1]{kojima_large_2023}. \textit{Analogical Prompting} instructs LLMs to create examples that can improve output quality using in-context learning \citep{yasunaga_large_2024}. \textit{Thread-of-Thought (ThoT)} reviews several prompting templates for efficiency, with the following instruction rated best: \enquote{Walk me through this context in manageable parts step by step, summarizing and analyzing as we go} \citep{zhou_thread_2023}. \textit{Plan-and-Solve} builds on the previously introduced Chain-of-Thought Zero-shot, but instead uses: \enquote{Let's first understand the problem and devise a plan to solve the problem. Then, let's carry out the plan and solve the problem step by step} \citep[pp.~3--4]{wang_plan-and-solve_2023}. \textit{Self-Consistency} uses Chain-of-Thought, but executes a prompt several times and decides, for example, in favor of the result whose solution was mentioned most frequently \citep[pp.~1--2]{wang_self-consistency_2023}. \textit{Tree-of-Thoughts} (ToT) also extends the Chain-of-Thought approach by following individual steps such as thought processes separately \citep[pp.~1--2]{yao_tree_2023}. \textit{Automatic Prompt Engineer} (APE) presents a system with which a prompt is selected from a set that leads to a high output quality. The following prompt was rated well in an evaluation: \enquote{Let's work this out in a step-by-step way to be sure we have the right answer} \citep{zhou_large_2023}.\\
Another noteworthy prompting technique is \textit{Self-Refine}, which uses an LLM to improve a result through feedback until a termination condition is reached \citep[pp.~1--2]{madaan_self-refine_2023}.
\section{Mapping Identified Techniques to the Prompt Canvas}
This chapter focuses on synthesizing the insights gathered from the literature review to populate the Prompt Canvas with relevant, evidence-based elements. It identifies key techniques in prompt engineering, aligning them with the structured sections of the canvas. Based on a user-centered design focusing on understanding the users' needs, the canvas consists of four categories, each containing a distinct aspect of a prompt: \textit{Persona/Role and Target Audience}, \textit{Goal and Step-by-Step}, \textit{Context and References}, \textit{Format and Tonality}. These categories align with the natural flow of information processing, from establishing the setting (persona and audience) to defining the task (goal and steps), providing the necessary background (context and references), and finally specifying the desired output (format and tone).
\subsection{Persona/Role and Target Audience}
Defining a specific persona or role helps in tailoring the language model's perspective, ensuring that the response aligns with the expected expertise or viewpoint. Identifying the target audience ensures that the content is appropriate for the intended recipients, considering their knowledge level and interests. This category is essential because it sets the foundation for the model's voice and the direction of the response, making it more relevant and engaging for the user.\\
This element was derived from recurring discussions in the literature about role-based prompting and user-centered design. Studies by \citet{braun_can_2024} highlighted the value of assigning roles to guide the model’s tone and specificity. \citet{sasson_lazovsky_art_2024} further emphasized the importance of personas in enhancing creative inquiry. These insights underscored the need to include a dedicated section on tailoring prompts to roles and audience characteristics.
\subsection{Task/Intent and Step-by-Step}
Clearly articulating the goal provides the language model with a specific objective, enhancing the focus and purpose of the response. Breaking down the goal into step-by-step instructions or questions guides the model through complex tasks or explanations systematically. This category justifies its inclusion by emphasizing the importance of precision and clarity in prompts, which directly impacts the quality and usefulness of the output.\\
Classified by \citet{braun_can_2024}, \citet{sahoo_systematic_2024} and \citet{sasson_lazovsky_art_2024}  as a distinct prompting category, Chain-of-Thought prompting techniques decompose a task step-by-step and enhance thereby the model's reasoning capabilities on complex problems. The Chain-of-Thought prompting technique can be used with both Zero-shot and Few-shot concepts. By structuring tasks incrementally, the model produces outputs that are both coherent and logically organized. Furthermore, this category facilitates creative inquiry; as \citet{sasson_lazovsky_art_2024} emphasize, clearly defining intent in prompts is essential for open-ended or exploratory tasks.
\subsection{Context and References}
Providing context and relevant references equips the language model with necessary background information, reducing ambiguity and enhancing the accuracy of the response. This category acknowledges that AI models rely heavily on the input prompt for context, and without it, the responses may be generic or off-target. Including references also allows the model to incorporate specific data or adhere to particular frameworks, which is vital in academic or professional settings.\\
This element was selected to address the frequent recommendation to provide situational and contextual information in prompts. \citet{braun_can_2024} stressed the importance of embedding contextual details to enhance output reliability and \citet{schulhoff_prompt_2024} suggested incorporating external references or historical data into prompts for guidance. Linking prompts to prior decisions, documents, or reports enhances contextual richness and ensures outputs reflect critical dependencies \citep{sasson_lazovsky_art_2024}. By integrating these elements, practitioners can craft prompts that are both informative and grounded in factual context.
\subsection{Output/Format and Tonality}
Specifying the desired format and tone ensures that the response meets stylistic and structural expectations. Whether the output should be in the form of a report, a list, or an informal explanation, and whether the tone should be formal, friendly, or neutral, this category guides the model in delivering content that is not only informative but also appropriately presented. This consideration is crucial for aligning the response with the conventions of the intended medium or genre.\\
This category emerged from the emphasis in the literature on aligning the model’s outputs with specific user requirements and communication contexts. Techniques like output specification and refinement, discussed in \citet{sahoo_systematic_2024}, are critical for aligning the model's output with user needs. \citet{braun_can_2024} highlighted specifying output formats to meet technical or domain-specific needs. Directing the model to produce responses in specific formats, such as tables, markdown, or code, ensures that outputs meet those requirements. Tonality customization and aligning tone with organizational branding to maintain consistency across communication outputs further validated the need to include this aspect in the Prompt Canvas. Also, it is of use to specify tone attributes like luxury, authority, or informality, depending on the target audience or purpose.\\ \\
By mapping the identified techniques to the Prompt Canvas, the foundational aspects of a prompt from defining personas to output refinement are systematically addressed. The canvas simplifies the application of complex techniques, making them more approachable for practitioners.\\
In addition to its primary elements, the integration of Techniques and Tooling categories serves to enhance the canvas by offering deeper technical insights and practical support. These categories focus on further techniques and the tools available to implement them.
\subsection{Recommended Techniques}
This category within the Prompt Canvas emphasizes the application of further strategies to refine and optimize prompts. These techniques enrich the Prompt Canvas by offering a diverse set of strategies to address varying tasks and contexts. Practitioners can draw from this toolbox to adapt their prompts to specific challenges.
\begin{description}[leftmargin=0cm, labelindent=0cm, itemsep=0.5em]
  \item[\textbf{Iterative Optimization}] \citet{sahoo_systematic_2024} and \citet{schulhoff_prompt_2024} present iterative refinement, through prompting techniques, as a crucial approach for improving prompts. This involves adjusting and testing prompts in a feedback loop to enhance their effectiveness. Iterative optimization allows practitioners to fine-tune prompts based on model responses, ensuring greater alignment with task objectives.
   \item[\textbf{Placeholders and Delimiters}] Placeholders act as flexible components that can be replaced with context-specific information, while delimiters help segment instructions, improving clarity and reducing ambiguity. Both can be used to create dynamic and adaptable prompts (cf. \citealt{white_prompt_2023}).
  \item[\textbf{Prompt Generator}] LLMs can also help to generate and refine prompts, making AI communication more effective. They assist in crafting precise instructions and optimizing existing prompts for better results.
  \item[\textbf{Chain-of-Thought Reasoning}] Chain-of-Thought encourages step-by-step reasoning in model outputs. By embedding sequential logic into prompts, practitioners can enhance the model’s ability to solve complex problems and provide coherent explanations.
  \item[\textbf{Tree-of-Thoughts Exploration}] Building on Chain-of-Thought methods, Tree-of-Thoughts prompting allows the model to explore multiple perspectives or solutions simultaneously. This technique is particularly valuable for tasks requiring diverse viewpoints or creative problem-solving.
  \item[\textbf{Emotion Prompting}] This technique involves appending emotional phrases to the end of a prompt to enhance the model's empathetic engagement \citep[p.~2]{li_large_2023}.
  \item[\textbf{Rephrase and Respond / Re-Reading}] As outlined by \citet[pp.~9--10]{deng_rephrase_2024} and \citet{xu_re-reading_2024}, these techniques have been shown to enhance reasoning performance.
  \item[\textbf{Adjusting Hyperparameters}] The ability to adjust hyperparameters such as \textit{temperature}, \textit{top-p}, \textit{frequency penalty}, and \textit{presence penalty} within the prompt itself is very helpful for controlling the diversity, creativity, and focus of the model’s outputs.
\end{description}
\subsection{Tooling}
The Tooling category offers practical support for designing and applying prompts efficiently. Tools and platforms simplify workflows, enhance accessibility, and enable the scalable deployment of prompt engineering techniques.
\begin{description}[leftmargin=0cm, labelindent=0cm, itemsep=0.5em]
  \item[\textbf{LLM Apps}] Apps like the ChatGPT App enable faster prompt creation through voice input, making interactions more efficient and accessible. This feature reduces typing effort, enhances usability on-the-go, and supports diverse users, streamlining the prompt engineering process for dynamic or time-sensitive tasks.
  \item[\textbf{Prompting Platforms}] Platforms like PromptPerfect allow users to design, test, and optimize prompts interactively. These tools often include analytics for assessing prompt performance and making informed adjustments.
  \item[\textbf{Prompt Libraries}] Pre-designed templates and reusable prompts, discussed by \citet{white_prompt_2023}, provide a valuable starting point for practitioners. Libraries save time and ensure consistency by offering solutions for common tasks. Some platforms either offer prompts for purchase (e.g., PromptBase), while others focus on sharing prompts for free (e.g., PromptHero).
  \item[\textbf{Browser Extensions}] Providing direct integration into web clients, browser extensions, like Text Blaze and Prompt Perfect, allow users to experiment with prompts in real-time on websites.
  \item[\textbf{LLM Arenas}] LLM Arenas, like Chatbot Arena, offer platforms to test and compare AI models, providing insights into their performance and capabilities. These arenas help users refine prompts and stay updated with the latest advancements in LLM technology.
  \item[\textbf{Custom GPTs for Specific Purposes}] \citet{Chen2024} mention GPTs as plugins in ChatGPT. Customized GPTs such as Prompt Perfect or ScholarGPT are tailored LLMs optimized for specialized applications or industries. These customized versions are also able to leverage additional data through Application Programming Interfaces (APIs) or are given additional context through text or PDFs for specific objectives, making them highly effective for specialized tasks.
  \item[\textbf{Customized LLMs and company-wide use}] Developing company-specific custom GPTs takes customization a step further by integrating organizational knowledge, values, and workflows into a LLM. These models have been given additional context or are even fine-tuned on internal data, leveraging documents and APIs, and are primed with internal prompts to ensure alignment with company standards and improve operational efficiency. Additionally, some LLM providers offer a sandboxed environment for enterprises, ensuring that entered data will not be used to train future publicly available models.
  \item[\textbf{Integration of LLMs via API into application systems}] APIs facilitate seamless integration of LLMs into existing systems, enabling automated prompt generation and application.
\end{description}
\section{Limitations, Outlook and Conclusion}
This chapter outlines the limitations of the current study, explores potential future directions for research and application, and concludes by emphasizing the significance of the Prompt Canvas as a foundational tool for the evolving field of prompt engineering. It provides a critical reflection on the scope of the work, its adaptability to emerging trends, and its role in bridging research and practice.
\subsection{Limitations}
As prompt engineering is not a one-size-fits-all discipline, different tasks and domains may require tailored approaches and techniques. Yet, a canvas can be easily customized to include domain-specific elements, such as ethical considerations for healthcare or creative constraints for marketing. This adaptability ensures that the canvas remains relevant and useful across diverse use cases. The modular structure allows practitioners to customize techniques for specific tasks or domains, improving relevance and scalability. \\
The effectiveness of this canvas requires validation through both quantitative and qualitative research methodologies. Recognizing the strong demand for a practical guide in the field of prompt engineering, this publication aims to serve as a starting point to initiate and foster discussion on the topic. Additionally, a research design to evaluate the utility of the canvas is already under development.\\
This work focuses primarily on text-to-text modalities. Although this modality should already cover a wide range of applications, there are other modalities such as image, audio, video or image-as-text (cf. \citealt{schulhoff_prompt_2024}) that are not highlighted in this study. At the same time, many techniques mentioned above are not designed exclusively for the text-to-text modality, e.g. iterative prompting. Furthermore, this work focused primarily on the design of individual prompts. Prompting techniques that use agents were thematically separated out in the analysis and synthesis. It is assumed that they can play another important role in further improving the output quality. At the same time, this work focused on findings for users of LLMs in the private and business environment. Finally, it is important to emphasize that this work does not explore potential risks associated with the use of LLMs. These risks include biases, handling of sensitive information, copyright violations, or the significant consumption of resources.
\subsection{Outlook}
The Prompt Canvas serves as a foundational tool, offering a shared framework for the field of prompt engineering. It is intended not only for practical application but also to foster dialogue about which techniques are most relevant and sustainable. By doing so, the canvas encourages discussion and guides research in evaluating whether emerging developments should be incorporated into its framework. Given the dynamic and rapidly evolving nature of the discipline, it is important to view the Prompt Canvas not as a static product but as a living document that reflects the current state of practice. For instance, if prompting techniques are more deeply integrated into LLMs in the future through prompt tuning and automated prompts, one could argue that some prompt techniques may become less important. Advancing models, such as OpenAI's o1 model series, already incorporate the Chain-of-Thought technique, enabling it to perform complex reasoning by generating intermediate steps before arriving at a final answer.
\subsection{Conclusion}
This paper introduces the Prompt Canvas as a unified framework aimed at consolidating the diverse and fragmented techniques of prompt engineering into an accessible and practical tool for practitioners. Grounded in an extensive literature review and informed by established methodologies, the Prompt Canvas addresses a need for a comprehensive and systematic approach to designing effective prompts for large language models. By mapping key techniques, such as role-based prompting, Chain-of-Thought reasoning, and iterative refinement, onto a structured canvas, this work provides a valuable resource that bridges the gap between academic research and practical application. Future research is encouraged to expand the framework to address these evolving challenges, ensuring its continued relevance and utility across diverse domains.
\bibliographystyle{plainnat} % alphabetical order by author

\clearpage
\appendix
\section{Appendix}
% tab:database_and_keyterms
\begin{table}[H]
\caption{Identified databases with respective search terms.}
%\vspace{0.1in}
\begin{tabularx}{\textwidth}{lX}
\toprule
\textbf{Database} & \textbf{Search term} \\
\midrule
\textbf{AIS eLibrary} & 
\textit{1)}~\enquote{prompt-engineering}; \textit{2)}~\enquote{prompt engineering}; \textit{3)}~\enquote{prompt techniques}; \textit{4)}~\enquote{prompt designs}; \textit{5)}~\enquote{prompt design}; \textit{6)}~\enquote{prompt patterns}; \textit{7)}~\enquote{prompt pattern}; \textit{8)}~\enquote{prompt strategies}; \textit{9)}~\enquote{prompt strategy}; \textit{10)}~\enquote{prompt methods}\\ &
\textit{The search in the AIS eLibrary was carried out individually for each term, as the website did not support OR operators.}\\
\addlinespace
\textbf{ACM Digital Library} & 
\texttt{Title: ("prompt-engineering" OR "prompt engineering" OR "prompt techniques" OR "prompt designs" OR "prompt design" OR "prompt patterns" OR "prompt pattern" OR "prompt strategies" OR "prompt strategy" OR "prompt methods")}\\
\addlinespace
\textbf{IEEE Xplore} & 
\texttt{("Document Title":prompt-engineering) OR ("Document Title":prompt engineering) OR ("Document Title":prompt techniques) OR ("Document Title":prompt designs) OR ("Document Title":prompt design) OR ("Document Title":prompt patterns) OR ("Document Title":prompt pattern) OR ("Document Title":prompt strategies) OR ("Document Title":prompt strategy) OR ("Document Title":prompt methods")}\\
\addlinespace
\textbf{Scopus} & 
\texttt{(TITLE("prompt-engineering") OR TITLE("prompt engineering") OR TITLE("prompt techniques") OR TITLE("prompt designs") OR TITLE("prompt design") OR TITLE("prompt patterns") OR TITLE("prompt pattern") OR TITLE("prompt strategies") OR TITLE("prompt strategy") OR TITLE("prompt methods"))}\\
\addlinespace
\textbf{arXiv} & 
\texttt{"prompt-engineering" OR "prompt engineering" OR "prompt techniques" OR "prompt designs" OR "prompt design" OR "prompt patterns" OR "prompt pattern" OR "prompt strategies" OR "prompt strategy" OR "prompt methods"}\\
\bottomrule
\end{tabularx}
\label{tab:database_and_keyterms}
\end{table}
\end{document}